\def\year{2022}\relax
\documentclass[letterpaper]{article} 
\usepackage{aaai22}  
\usepackage{times}  
\usepackage{helvet} 
\usepackage{courier}  
\usepackage[hyphens]{url}  
\usepackage{graphicx} 
\usepackage{graphicx}  
\usepackage{subfigure}
\usepackage{amsmath,amsfonts}
\usepackage{booktabs}
\usepackage{xcolor}
\usepackage{amsthm}
\usepackage{mathtools}
\DeclarePairedDelimiterX{\inp}[2]{\langle}{\rangle}{#1, #2}
\newtheorem{theorem}{Proposition}
\newtheorem{lemma}[theorem]{Lemma}

\title{Nonparametric Topological Layers in Neural Networks}
\author{Dongfang Zhao\\ University of Nevada, Reno
}
 
\begin{document}

\maketitle

\begin{abstract}
Various topological techniques and tools have been applied to neural networks in terms of network complexity, explainability, and performance. 
One fundamental assumption of this line of research is the existence of a global (Euclidean) coordinate system upon which the topological layer is constructed.
Despite promising results, such a \textit{topologization} method has yet to be widely adopted because the parametrization of a topologization layer takes a considerable amount of time and more importantly, lacks a theoretical foundation without which the performance of the neural network only achieves suboptimal performance.
This paper proposes a learnable topological layer for neural networks without requiring a Euclidean space;
Instead, the proposed construction requires nothing more than a general metric space except for an inner product, i.e., a Hilbert space.
Accordingly, the according parametrization for the proposed topological layer is free of user-specified hyperparameters,
which precludes the costly parametrization stage and the corresponding possibility of suboptimal networks.
\end{abstract}

\section{Introduction}

\subsection{Background}

Topological techniques and optimizations have been proposed for neural networks from various aspects such as network complexity~\cite{mmoor_icml20}, explainability~\cite{gnait_jmrl20}, and performance~\cite{brieck_icml19}. 
As opposed to the conventional, geometrical standpoint for training a neural network, topological approaches are concerned with the \textit{qualitative} properties of the underlying data.
Generally speaking, a geometrical approach imposes the \textit{quantitative} metric between objects (e.g., Euclidean distance),
whereas a topologist is more interested in identifying the intrinsic property of the targeted objects.
Informally, an \textit{intrinsic} property refers to the property that is invariant under ``continuous'' deformation of the object's geometrical form.
It is evident to see the benefits of employing topological information in the applications of neural networks;
we give two examples in the following.

One of the most widely used topological invariants is \textit{connectedness}.
While connectedness is rigorously defined by topologists at arbitrary dimensions,
it suffices to understand connectedness in the conventional sense of graph theory (at least in $\mathbb{R}^2$):
As long as there is a path between any pair of vertices in the graph, 
the graph is \textit{connected}.\footnote{Mathematically, the connected graphs are called \textit{path-connected} or $0$-connected. We do not use such terms to avoid unnecessary confusion.}
Informally, a figure `8' (in $\mathbb{R}^2$) is in one piece (i.e., connected),
while an English letter `i' is not.
Moreover, we can reasonably assume that the handwritten `8' and `i' would retain the connectedness property,
i.e., both figures are under continuous deformation and can be distinguished by their connectedness.

A somewhat less obvious (topological) invariant is \textit{homology group}.
Roughly speaking, the homology group captures the cycles of the underlying object but the cycle is not a boundary of any region (i.e., subset) of the object.
A nice geometrical intuition of those homology groups is, therefore, \textit{holes}.
It is well known that these homology groups (up to homomorphism), or more specifically the ranks of those groups, do not change under continuous deformation.
Taking again the example of handwritten figures,
`8' has two 1-dimensional holes, `0' has one 1-dimensional hole, and `1' has none.

As stated in the beginning, incorporating topological invariants has drawn a lot of research interest, as illustrated in the above two examples.
The state-of-the-art topological extensions~\cite{tlacombe_ijcai21,mcarriere_icml21,kkim_nips20} of neural networks are built upon the so-called \textit{persistent topology}~\cite{hedels_book10},
which essentially determines those topological invariants from a series of parametrization over the underlying object.

\subsection{Motivation and Challenges}

The common assumption of existing works for employing topology into neural networks is the existence of a global (Euclidean) coordinate system,
upon which the layer of persistent topology is parametrized. 
However, the parameterization procedure is ad-hoc, 
and there yet exists a systematic manner for tuning the knobs of (hyper)parameters.
In fact, state-of-the-art methods for constructing a topological layer for neural networks adopt a polynomial~\cite{rgabriel_aistat20} or exponential~\cite{chofer_nips17} expression over the infinite $\mathbb{R}$ field,
making the parameter space \textit{infinite}.
The parametrization procedure, therefore, inevitably leads to both high preprocessing costs and there is no guarantee that the resulting model is optimal.
We thus argue that there is a pressing need for a more efficient approach to integrating topology into neural networks.

This paper proposes a \textit{nonparametric} construction of a learnable topological layer.
The rationale of our proposal lies at the root cause of the parameterization challenge: 
The Euclidean space is an \textit{overly} rich topological (metric) space in the sense that every single point can be globally identified by a sequence of real-valued coordinates (which are infinite).
If we can migrate the parametrization of those points (in terms of Euclidean coordinates) into some intrinsic components that are implicitly characterized by the application, 
we are then free of the turning tasks of the infinite parameter space.
In \textit{functional analysis}, which studies the relationship between functions rather than values, 
some spaces that are more ``abstract'' than the Euclidean space, 
such as Banach spaces~\cite{pbub_jmlr15} and Hilbert spaces~\cite{bspipe_jmlr10} (e.g., reproducing kernel Hilbert space, or RKHS),
have drawn peculiar interests as they allow us to quantify the relationships among functions.
This work is inspired by the above works where the learnable topological layers can be embedded in a Hilbert space.

The technical challenge of employing nonparametric topological layers in a functional space is twofold.
First, because the topological layer is constructed without a global Euclidean coordinate system, we need to ensure (i.e., prove) that the layer is \textit{differentible} such that the other metrics (e.g., gradient)  are well-defined in the neural network applications.
Second, it is unclear how to construct meaningful subsystems/criteria (e.g., loss function) using a limited set of metrics (e.g., norms and inner products in a Hilbert space).
To our knowledge, there is no prior work investigating a nonparametric method for incorporating topological layers into neural networks.  

\subsection{Contributions}

This paper presents the construction of a learnable topological layer embedded in a Hilbert space.
To the best of our knowledge, this is the first learnable topological layer constructed using only inner products,
which eliminates the high cost of the topological layer's parametrization.
We prove that the proposed metrics are continuous and more importantly, that the derived loss function used in a gradient-based learning procedure is differentiable.
We experimentally demonstrate that the proposed nonparametric topological layer achieves competitive performance compared to the state-of-the-art parametrized methods.

\section{Methodology}

\subsection{Preliminaries of Topology and Analysis}

We briefly review concepts and facts (without proof) that will be used later.
Readers who are familiar with topological spaces and functional analysis can skip this subsection.

A \textit{topology} of a set $S$ is a collection of subsets of $S$, denoted $\mathcal{T}$.
One example topology of $S$ is then the power set of $S$, $\mathcal{P}(S)$, 
which consists of all the possible $2^{|S|}$ subsets of $S$.
This is also called the \textit{discrete topology} of $S$.
The pair $(S, \mathcal{T})$ is called a \textit{topological space};
in practice, we usually adopt more information to characterize the topological space with additional properties.
Each of the subsets $U$ from $\mathcal{T}$ is called an \textit{open set}, and the complement set $S \setminus U$ is a \textit{closed set} by definition.
A function $g$ from space $X$ to $Y$ is called \textit{continuous} if:
$\forall v$ is an open set in $Y$, then $g^{-1}(v)$ is an open set in $X$.
The composition of two continuous functions is also continuous.
If both $g$ and $g^{-1}$ are continuous and bijective, we call $g$ a \textit{homeomorphism}.
Since a homeomorphism is defined on open and closed sets,
two topological spaces are \textit{equivalent up to homeomorphism} if the latter exists.

A \textit{metric space} is a set $S$ of points where a function $d$ is defined for every pair of points, $d: S \times S \rightarrow \mathbb{R^*}$, where $\mathbb{R^*}$ denotes the set of non-negative real numbers and $d$ satisfies the following properties (axioms): 
$\forall x,y,z \in S$, $d(x,x) = 0$, $d(x,y) = d(y,x)$, and $d(x,y) \leq d(x,z) + d(z,y)$.
Obviously, the Euclidean space turns out to be a metric space with the $d$ function defined as the Euclidean distance $\sqrt{ \sum^n_{i=1} (x_i - y_i)^2}$ in $\mathbb{R}^n$, where $x_i$ and $y_i$ are the coordinates of two points.
In the context of neural network applications, a metric space is always \textit{complete} because the set of real numbers $\mathbb{R}$ is a complete metric space,
which implies that there is no ``gap'' in a convergent sequence of points.\footnote{Technically, this means that any Cauchy sequence converges in the space.
A rigorous proof is beyond the scope of this paper and can be found in any elementary textbook of functional analysis.}

A \textit{normed space} is a metric space where a norm function, usually denoted $\Vert \cdot \Vert$, is well defined and satisfies the following properties:
$\forall x,y \in S, \forall \lambda \in \mathbb{R}$,\footnote{Technically, $\lambda$ could be an element in any other \textit{field} $\mathbb{F}$. We restrict our discussion to the real numbers $\mathbb{R}$ (which is also a field) in the context of neural network applications.}
$\Vert x + y \Vert \leq \Vert x \Vert + \Vert y \Vert$, 
$\Vert \lambda x \Vert = |\lambda| \Vert x \Vert$, and
$\Vert x \Vert = 0 \Leftrightarrow x = 0$.
A complete, normed space is also called a \textit{Banach space}.
Informally, the norm function $\Vert \cdot \Vert$ measures the ``length'' of a point from the ``origin''.
Once again, it is evident to verify the Euclidean space $\mathbb{R}^n$ is a normed space with $\Vert \cdot \Vert$ defined as the Euclidean distance between a vector and $\mathbf{0} \in \mathbb{R}^n$.

An \textit{inner product space} is a metric space associated with a function denoted $\langle \cdot , \cdot \rangle$ from a pair of points to a value,
i.e., $\langle \cdot , \cdot \rangle: S \times S \rightarrow \mathbb{R}$ with the following properties:
$\forall x, y, z \in S, \forall \lambda \in \mathbb{R}$,
$\langle x, y + z \rangle = \langle x, y \rangle + \langle x, z \rangle$,
$\langle x, \lambda y \rangle = \lambda \langle x, y \rangle$,
$\langle x, y \rangle = \langle y, x \rangle$,
$\langle x, x \rangle \ge 0$, and
$\langle x, x \rangle = 0 \Leftrightarrow x = 0$.
Now, if we define a norm on an inner product space $X$ with 
$\Vert x \Vert = \sqrt{\langle x, x \rangle}$,
$X$ turns to be a normed space by verifying that $\Vert x \Vert$ indeed satisfies the three conditions in the last paragraph.
That is, we induce a single-variable norm function from the inner products with a pair of identical points.
As before, in the context of machine learning applications, all numbers are in $\mathbb{R}$ and thus all inner product spaces under consideration are complete,
which means that inner product spaces are Banach spaces by definition.
The important point here is that an inner product space provides not only the norm (i.e., ``length'') of points but also the relationship among the points.
Specifically, when the inner product between two points are zero,
i.e., $\langle x, y \rangle = 0$,
we call $x$ and $y$ are \textit{orthogonal} denoted by $x \bot y$.
A \textit{Hilbert space} is an inner product space and complete as a metric space, and the calculation of inner products turn to be the integral of functions ($f$ and $g$) over their domain $X$: 
$\int_{X} f(x) g(x) dx$.
For instance, if $X = [0,1] \times [0,1]$, $x = (x_1, x_2)$, $f(x) = \frac{x_1 + x_2}{2}$, $g(x) = \frac{x_1 - x_2}{2}$, then $\inp{f}{g} = \frac{1}{4}\iint \left( x_1^2 - x_2^2 \right) dx_1 dx_2 = 0$, i.e, $f \bot g$.

\subsection{Case Study of Persistent Topology}

\begin{figure}[htbp]
	\includegraphics[width=\linewidth]{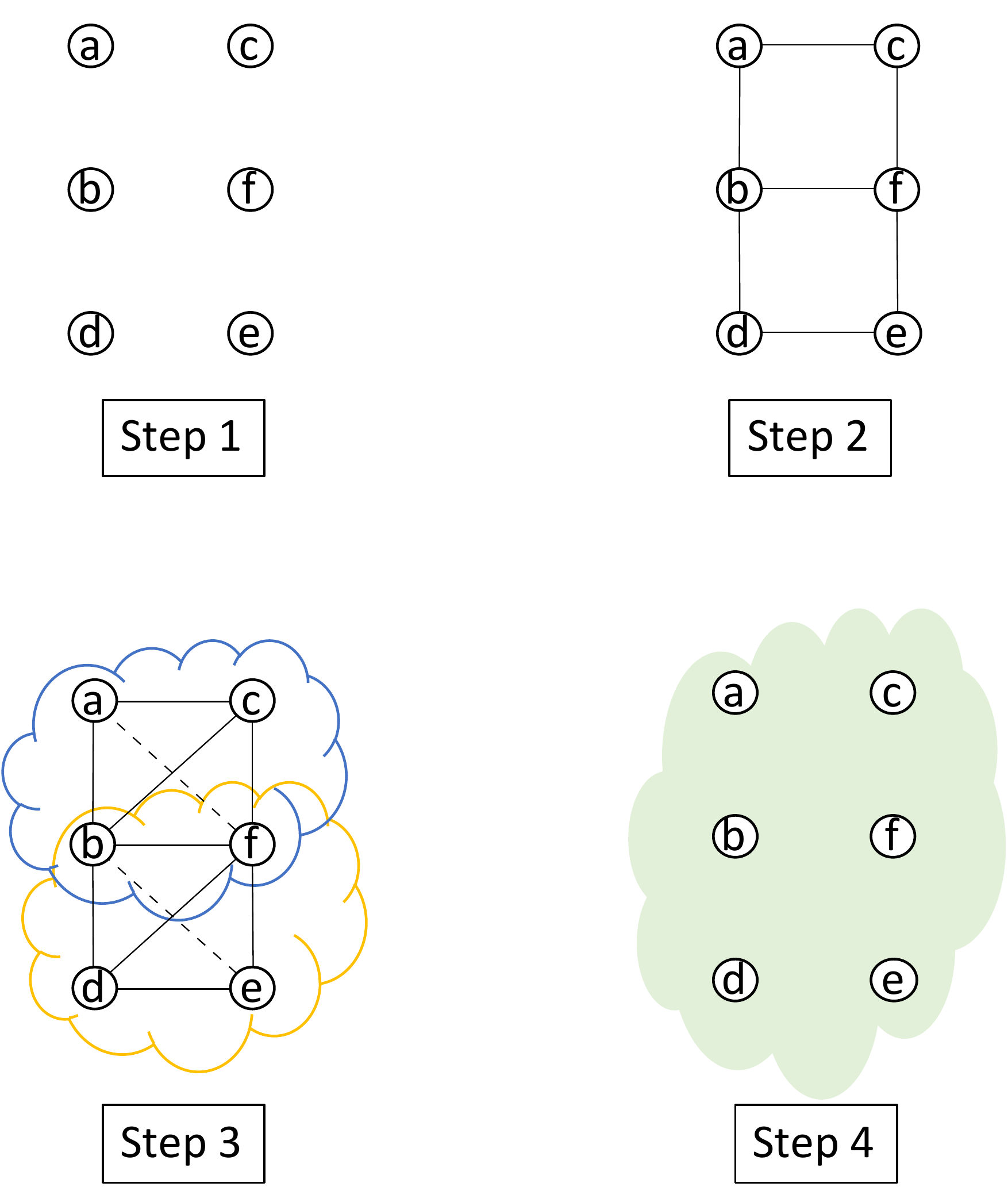}
    \caption{Four (out of many) simplicial complexes for six vertices of figure `8'.
    }
    \label{fig:eight_complex}
\end{figure}

Before formally describing the nonparametric layer,
we illustrate the procedure of feeding topological features into a learnable layer in the context of recognizing a hypothetical figure '8' that is composed of six vertices and a series of growing diameters connecting the vertices.
In Figure~\ref{fig:eight_complex}, we show four possible \textit{complexes}, denoted \textbf{Step $\mathbf{i}$}.
A (simplicial) \textit{complex} is defined as a union of \textit{simplices} (singular: simplex);
a \textit{simplex} is defined as an open set all of whose subsets are open sets as well.
An $n$-dimensional simplex has $n+1$ points, often called an $n$-simplex for short.
The step numbers imply the sequence of procedures we will discuss soon;
for now, we can treat these four complexes independently.
We assume that the (Euclidean) distance between every pair of ``close'' vertices is $\delta$:
\[
\overline{ba} = \overline{ac} = \overline{cf} = \overline{fe} = \overline{ed} = \overline{db} = \overline{bf} = \delta.
\]
Therefore, we have
\[
\overline{af} = \overline{cb} = \overline{be} = \overline{fd} = \sqrt{2}\delta,
\]
and 
\[
\overline{ae} = \overline{cd} = \sqrt{5}\delta.
\]

\textbf{\textbf{Step 1.}}
This is a trivial complex where there is no edge in it. The complex has six 0-dimensional simplices and no higher-dimensional simplices.

\textbf{\textbf{Step 2.}}
In this complex, we have seven edges, or seven 1-simplices, in addition to the six vertices.
There are no higher-dimensional simplices, as no ``plane'' is visible.
Notably, there are two ``independent'' loops, 
namely $\widehat{acfba}$ and $\widehat{bfedb}$;
the outline $\widehat{acfedba}$ is less interesting because it can be composed by the prior two shorter loops if we count the addition of two simplex in $\mathbb{F}_2$:
$\widehat{bf} + \widehat{bf} \equiv 0 \texttt{ mod } 2$  (see~\cite{jmunkres_book84} for more detail of \textit{modulo-2 homology}).

\textbf{\textbf{Step 3.}}
There are two simplices, indicated by two shallow clouds, respectively.
Each of the two simplices is three-dimensional, or a tetrahedron, with the sixth edge ``under the paper''.
The two simplices share a one-dimensional simplex, or a \textit{face}, i.e., edge $\widehat{bf}$.
Because $\widehat{bf}$ is also a (lower-dimensional) simplex, or \textit{proper face}, 
the entire simplicial complex is valid.

\textbf{\textbf{Step 4.}}
This is another trivial complex where all the six vertices belong to the same simplex. 
As a consequence, the complex is itself a six-dimensional simplex,
which cannot be visualized.
We simply denote the simplex with a solid cloud.
Technically, this complex is \textit{contractible},
i.e., topologically equal to a single point up to homeomorphism.

\begin{figure}[htbp]
	\includegraphics[width=\linewidth]{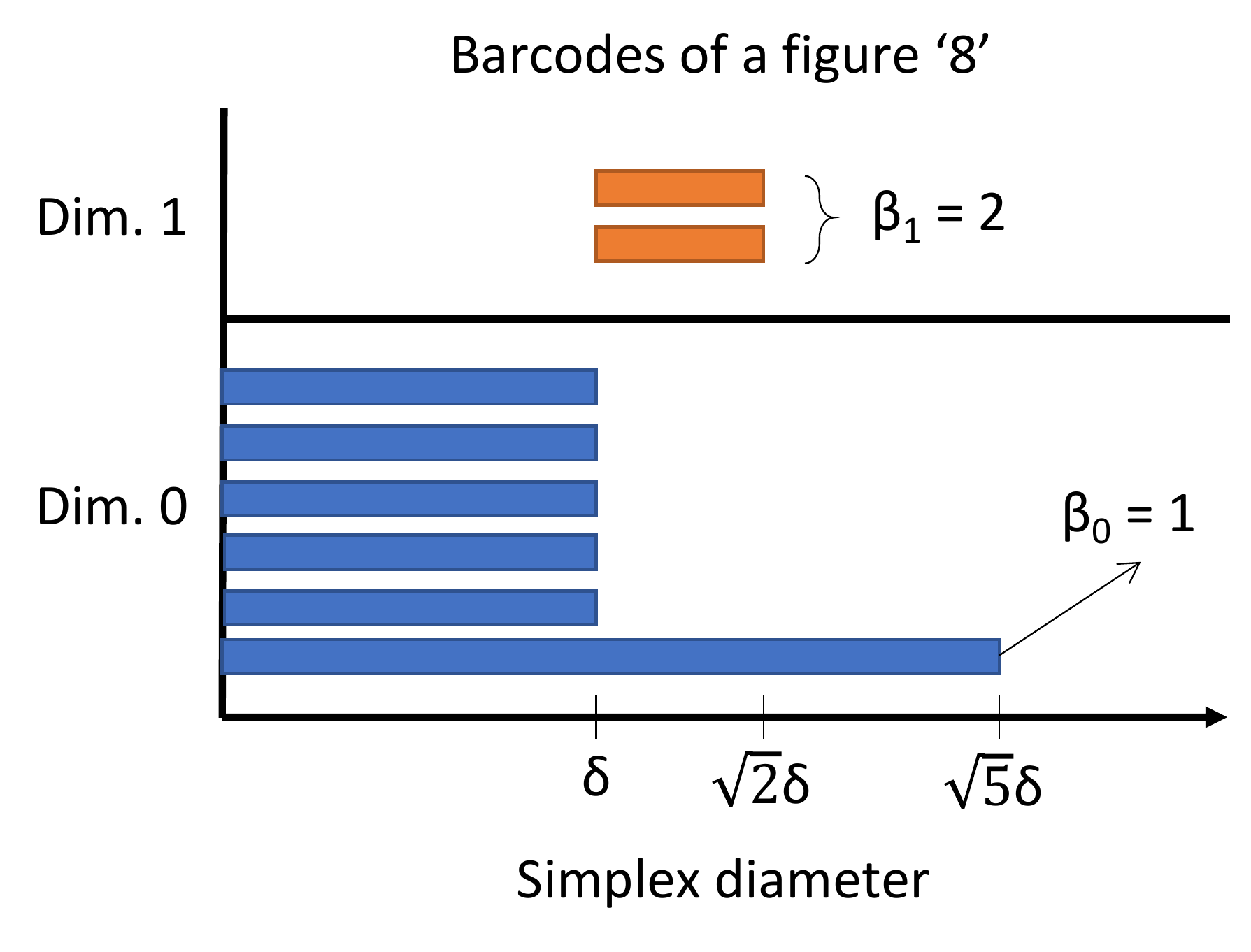}
    \caption{The barcodes of figure `8' used to feed a neural network for classification. Each bar starts at the \textit{birth time} (the left endpoint) and ends at the \textit{death time} (the right endpoint)}
    \label{fig:eight_barcodes}
\end{figure}

The 0- and 1-dimensional topological invariants are then represented in a diagram with $x$-axis for diameters and $y$-axis for distinct dimensions,
as shown in Figure~\ref{fig:eight_barcodes}.
Each of the barcodes represents a unique topological feature in respective dimensions.
Each of the four steps discussed above is considered an important ``milestone'' where the overall topology is changed.
Essentially, we want to (re-)calculate the algebraic invariant when new simplices appear (birth) or existing simplices disappear (death). 
Note that an existing simplex that will disappear can be thought of joining another simplex to form a new simplex;
What happens to five 0-dimensional bars at $\delta$ is that those six components appearing at the beginning of the procedure merge into a single component.
This single component continues to live till the end of the procedure---when the complex includes everything with a diameter of the diagonal line between vertices $a$ and $e$, $\sqrt{5}\delta$.
Looking back at $\delta$, we see two new 1-dimensional ``holes'' appear,
corresponding to the two rings in figure `8'. 
Both holes, however, disappear when the diameter reaches the length of the diagonal line between vertices $a$ and $c$;
at this point, the space encompassed by the path $\widehat{acfba}$ is fulfilled by the simplex and the ``hole'' is gone.
Exactly the same thing happens to the hole formed by $\widehat{bfedb}$.

Another note on barcodes is about their completeness:
the diagram of barcodes does not necessarily exhibit all the possible complexes.
For instance, in the above example, there is another milestone where the complex does change:
when the grid diameter increased to $2\delta$, 
$\sqrt{2}\delta < 2\delta < \sqrt{5}\delta$,
vertex $a$ and vertex $d$ can now be directly connected.
At this point, we have four distinct five-dimensional simplices centered vertices $a$, $c$, $e$, and $d$, respectively.
While this represents a distinct simplicial complex than the ones shown in Steps 1--4,
it exhibits no new information to the topological invariant, and therefore, we do not plot it in the barcodes of Figure~\ref{fig:eight_barcodes}.
The $i$-th barcode can also be represented with the length ($L_i = d_i - b_i$) and the mean value ($M_i = \frac{d_i+b_i}{2}$) with respect to the diameter,
both of which play an important role in the nonparametric layer that we will discuss in the next subsection.

\subsection{Constructing the Nonparametric Layer}

We initialize the topological layer with the norms of input data and update the node weights in a Hilbert space.
Similar to TopoSig~\cite{chofer_nips17}, the nonparametric topological layer (NTL) is prepended to the neural network.
The topological information used to feed into the NTL is also called the input's (topological) \textit{signature},
which is usually communicated through multisets (or, bag) with possibly duplicate values. 
In the following, we will be focusing on the construction of a nonparametric loss function in the Hilbert space.
Before that, we quickly review the status quo~\cite{rgabriel_aistat20} for adopting the topological signature in a \textit{polynomial} form at a specific dimension:
\[\displaystyle
\Delta F = - \sum_j (d_j - b_j)^p \times \left(\frac{d_j + b_j}{2}\right)^q,
\]
where $j$ is the index of a specific cycle (bar), $b_j$ and $d_j$ indicate the birth time and death time (see the examples in Fig.~\ref{fig:eight_barcodes}), and $p$ and $q$ are the weights for the lengths and the means, respectively.

Assuming that there are a total of $m$ cycles at dimension $i$.
Evidently, we have $m \le n$ because there cannot exist more than $n$ cycles with $n$ points.
We code the cycles with a pair of real numbers $(l, m) \in \mathbb{R}^2$.
Denote two points with vector $L_i$ and $M_i$, respectively, for dimensional $i$.
The question then becomes to identify a function $f$ such that:
\[
f: L \times M \rightarrow F^n.
\]

To guarantee that the layer is differentiable, as is required for updating the neural network structure, 
we need to show that the $f$ function is continuous.
Intuitively, when we think about two vectors with an inner product, a tiny movement of any of the two vectors should only lead to a similarly tiny change to the projected value.
We will formally prove this claim using point-set-topological machinery, as follows.

\begin{theorem}
The inner product over $L$ and $M$ of dimension $m$ is a continuous function with the codomain $F$ of dimension $n \ge m$.
\end{theorem}
\begin{proof}
Let $V$ be a $n$-dimensional open set in $F$.
Our goal is to show that $U = f^{-1}(V)$ is a pair of open sets in the $m$-dimensional space.
Note that an open set $O \in S$ is nothing but a subset with the following three axioms (see~\cite{jmunkres_book03} for introductory point-set topology):
\begin{itemize}
    \item [O1:] The union of any number of open sets is an open set;
    \item [O2:] The intersection of finite open sets is an open set; and
    \item [O3:] Both $S$ and $\emptyset$ are open sets.
\end{itemize}

Let $U_j = f^{-1}(V_j) = (U_j^1, U_j^2)$ denote the preimage of an arbitrary open set $V_j \in F$,
i.e., the pair of $m$-dimensional points.
By definition, an inner product of two $m$-dimensional points is the projection of one vector onto the other.
That is, $\inp{U_j^1}{U_j^2}$ is an open set.
We will show that $U_j^1$ is an open set; $U_j^2$ can be proven symmetrically.

For contradiction, suppose $U_j^1$ is not an open set.
Then, by O1, $U_j^1$ cannot be a union of open sets.
We impose the same direction of $U_j^1$ on $\inp{U_j^1}{U_j^2}$, denoted $U_j^{\alpha}$,
which is evidently an open set.
Let $U_j^{\beta} = U_j^{1} - U_j^{\alpha}$,
then $U_j^{\beta}$ cannot be an open set because otherwise $U_j^{1}$ would be an open set---a union of two open sets.
Note that $U_j^{\alpha}$ (an open set) and $U_j^{\beta}$ (a non-open set, or \textit{closed set}) are linear dependent.
It follows that a linear function can map between an open set and a closed set,
which is impossible because the limit points of a closed set (i.e., the boundary) cannot be linearly mapped to an open set.
We thus reach a contradiction and the claim is proved.
\end{proof}

As the next step, we need to aggregate each dimension's contribution.
In real-world neural network applications, such as recognizing MNIST, we want to consider characteristics from multiple dimensions like 0-dimensional \textit{single components} (e.g., one component for figure `7') and 1-dimensional \textit{holes} (e.g., two holes for figure `8').
Therefore, the overall \textit{nonparametric} loss function looks the following:
\[\displaystyle
\mathcal{F} = \Delta F^n \coloneqq \sum_{i=0}^n  (-1)^i \cdot (1 + dim(\Delta F_i)) \cdot  \langle L_i,  M_i \rangle,
\]
where $i$ indicates a specific dimension of the simplicial complex.
We then need to show that $\Delta F^n$ is differentiable. 
Note that we do not assume a Euclidean space or Euclidean coordinates,
therefore we will have to show that all the partial derivatives (i) exist and (ii) are continuous.
We sketch the proofs of both properties in the following two Lemmas.

\begin{lemma}\label{thm:part_exist}
Both $\frac{\partial}{\partial L}\mathcal{F}$ and $\frac{\partial}{\partial M}\mathcal{F}$ exist.
\end{lemma}

\begin{proof}
Recall that a partial derivative by definition is nothing but a directional derivative over the variables.
For $L$, we have
\[\displaystyle
\begin{split}
\frac{\partial}{\partial L}\mathcal{F} & = \lim_{h \rightarrow 0} \Bigg( \Big( \sum_i (-1)^i (1 + dim(\Delta F^n_i)] \langle L_i + h_i,  M_i \rangle 
\\ 
& \hspace{1cm} - \sum (-1)^i [1 + dim(\Delta F^n_i)) \langle L_i,  M_i \rangle \Big) / \Vert h \Vert \Bigg)\\
&= \sum_i (-1)^i (1 + dim(\Delta F^n_i)) \\
& \hspace{2cm} \times \lim_{h \rightarrow 0} \frac{\langle L_i + h_i,  M_i \rangle - \langle L_i,  M_i \rangle}{\Vert h \Vert} \\
&= \sum_i (-1)^i (1 + dim(\Delta F^n_i)) \cdot \lim_{h \rightarrow 0} \frac{\langle h_i,  M_i \rangle}{\Vert h \Vert} \\
& = \sum_i (-1)^i (1 + dim(\Delta F^n_i)) \cdot \lim_{e^i \rightarrow 0} \inp{e^i}{M_i} \\
& = \sum_i (-1)^i (1 + dim(\Delta F^n_i)) \cdot  \int_{B_{\epsilon}} e^i(x)M_i(x)dx,
\end{split}
\]
where $e^i = \frac{h_i}{\Vert h \Vert}$ indicates the (normalized) coordinates and $B_\epsilon$ indicates the small neighborhood around the point where the derivative is applied.
Intuitively, the factor $\int_{B_{\epsilon}} e^i(x)M_i(x)dx$ in the last equality represents the ``movement'' of the function at point $(L_0, M_0)$ in the direction $h$.\footnote{Mathematically speaking, it would be the external derivative of a vector field $\mathcal{H}_{(L_0, M_0)}[\mathcal{F}]$. We do not use such terms to avoid unnecessary confusions.}
Recall that our construction is in a Hilbert space,
which by definition means that there are no ``missing'' points in any sequence. 
Therefore, the last factor does not diverge and must exist:
$\int_{B_{\epsilon}} e^i(x)M_i(x)dx < \infty$. 
As a result, $\frac{\partial}{\partial L}\mathcal{F}$ exists, as desired.

The existence of $\frac{\partial}{\partial M}\mathcal{F}$ can be proved similarly.
\end{proof}

\begin{lemma}\label{thm:part_contin}
Both $\frac{\partial}{\partial L}\mathcal{F}$ and $\frac{\partial}{\partial M}\mathcal{F}$ are continuous.
\end{lemma}

\begin{proof}
It is sufficient to show that the difference between the inner product of the component sequences converges to zero given that each component converges to zero.
That is, we need to show 
\[
n \rightarrow \infty,  L_n \rightarrow L,  M_n \rightarrow M \Rightarrow \langle  L_n,  M_n \rangle \rightarrow \langle L, M \rangle.
\]
Equivalently, we claim the following limit
\[
|\langle  L_n,  M_n \rangle - \langle L, M \rangle| \rightarrow 0.
\]
First, we can compute the difference between two inner products as follows (recall the linearity property in inner products):
\[
\begin{split}
\langle  L_n,  M_n \rangle - \langle L, M \rangle &= \inp{ L_n-L}{ M_n} + \inp{L}{ M_n}\\
& \hspace{1cm} - (\inp{L}{M- M_n} + \inp{L}{ M_n}) \\
&= \inp{ L_n-L}{ M_n} + \inp{L}{M- M_n}.
\end{split}
\]
Then, by triangle inequality we have:
\[
\begin{split}
& \hspace{5mm} |\langle  L_n,  M_n \rangle - \langle L, M \rangle| \\
& \le |\inp{ L_n-L}{ M_n}| + |\inp{L}{M- M_n}| \\
& \le \sqrt{\inp{L_n-L}{L_n-L}}\sqrt{\inp{M_n}{M_n}} \\
& \hspace{1cm} + \sqrt{\inp{L}{L}}\sqrt{\inp{M-M_n}{M-M_n}} \\
& \rightarrow \sqrt{\inp{0}{0}}\sqrt{\inp{M_n}{M_n}} + \sqrt{\inp{L}{L}}\sqrt{\inp{0}{0}}\\
& = 0 \cdot \Vert M_n \Vert + \Vert L \Vert \cdot 0 \\
& = 0,
\end{split}
\]
where the last equality is because both norms $\Vert M_n \Vert$ and $\Vert L \Vert$ are finite in persistent homology.
\end{proof}

\begin{theorem}
$\mathcal{F}$ is differentiable.
\end{theorem}
\begin{proof}
The claim immediately follows Lemmas~\ref{thm:part_exist} and~\ref{thm:part_contin}.
\end{proof}

\section{Evaluation}
\label{sec:eval}

\subsection{Experimental Setup}\label{subsec:eval_setup}

\textbf{Testbed.}
All of our experiments are carried out on a 64-bit Ubuntu 20.04 LTS workstation.
The workstation has an Intel Core Quad i7-6820HQ CPU @ 2.70GHz, 64 GB memory, and a 1 TB Samsung SSD drive.
Our Anaconda environment has the following libraries installed:
Numpy~1.19.2, 
Pandas~1.1.3,
Python~3.8.5,
PyTorch~1.4.0,
and
Scipy~1.5.2.
Our C++ compiler is g++~9.3.0.

\textbf{Data sets.}
We use three data sets for experiments, all of which are publicly available: MNIST~\cite{mnist}, KMNIST~\cite{kmnist}, and FashionMNIST~\cite{fmnist}.
All of them comprise 28$\times$28 data points, 60,000 for training and 10,000 for testing.
We repeated all the experiments at least five times and observed unnoticeable differences;
for this reason, we will not report the error bars of the reported numbers in the remainder of this section.

\textbf{Network models.}
We use the same neural network model in the PyTorch implementation of MNIST~\cite{mnist}. 
The network has two convolution layers for channel numbers $1 \rightarrow 32 \rightarrow 64$ with kernel size 3.
The dropout rates are 0.25 and 0.5, respectively.
The dimensions of linear transformation are $9,216 \rightarrow 128 \rightarrow 10$.
A Rectified Linear Unit (ReLU) layer is applied after the first layer, and a LogSoftmax function is applied after the second layer.
We set the batch size as 100 and the epoch number as nine.
The model is trained with the entire training set (60,000) and evaluated with the entire test set (10,000) unless otherwise stated.
GPU is disabled in our experiments.

\subsection{Implementation}\label{subsec:impl}

We implemented the proposed approach atop the codebase of~\cite{rgabriel_aistat20}.
The core algebraic computation was implemented with C++, and the neural network-related routines are either inherited from PyTorch or implemented from scratch with Python.


To incorporate the proposed method into an end-to-end application,
we also implemented a module to binarize the pixels of input grey-level data such that the simplicial complexes are built upon a set of discrete points.
When evaluating various network models and parameters, all the input data are binarized for a fair comparison.

One caveat when implementing the methods is that the topologization might move a point out of the 28$\times$28 frame.
This did not happen at all for MNIST and started showing up for KMNIST.
Our current solution to this problem is to still include the off-scene point at the boundary of the frame such that we do not lose the topologization effect.

\subsection{Overall Performance}

\begin{figure}[htbp]
	\includegraphics[width=\linewidth]{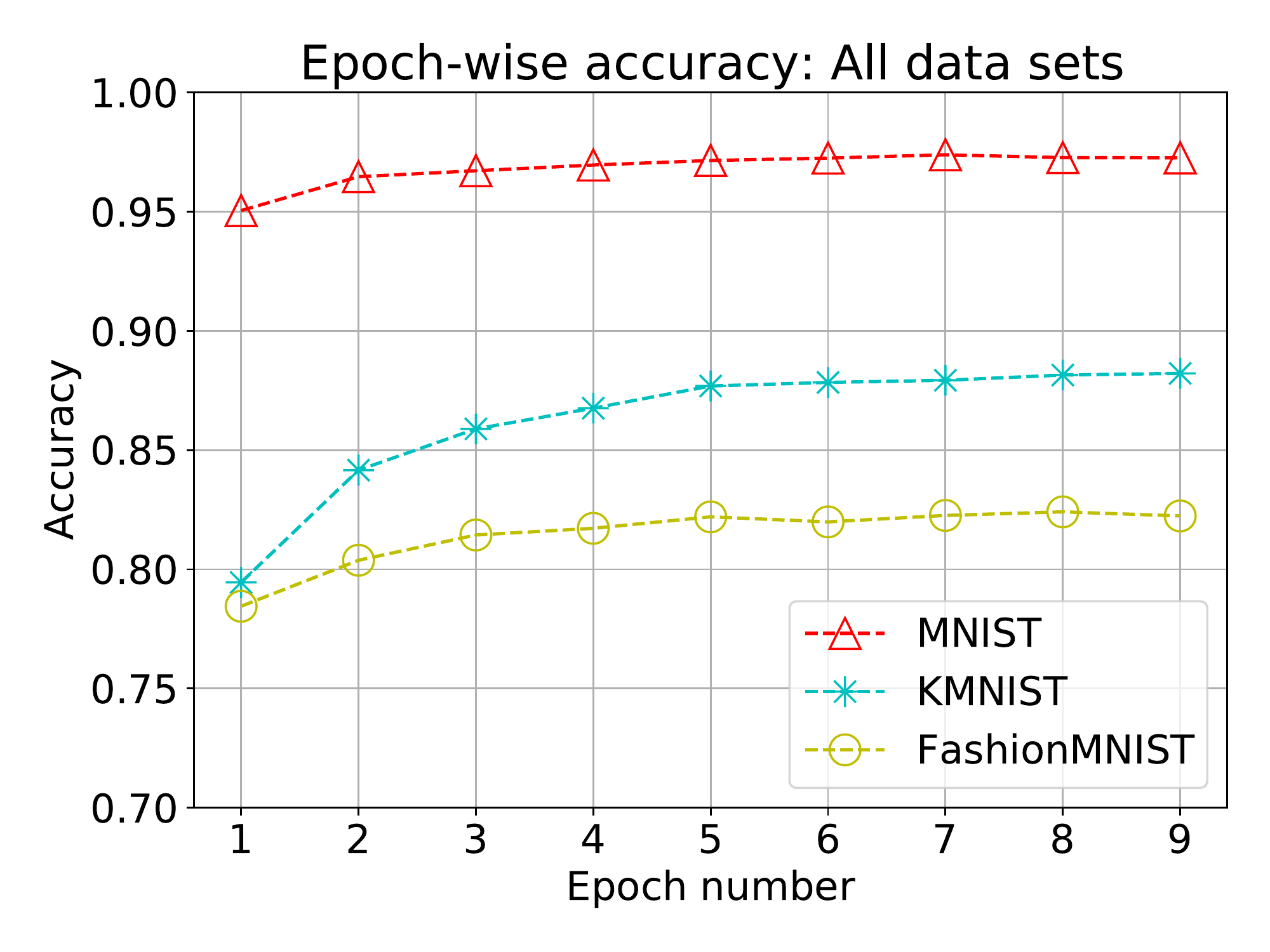}
	\caption{Accuracy at various epochs.}
	\label{fig:nn_accuracy_all_fullsize}
\end{figure}

We report the overall performance of the nonparametric topological layer in Figure~\ref{fig:nn_accuracy_all_fullsize}.
The accuracy after training at each epoch is plotted for all three data sets.
A common phenomenon is that the first five epochs represent the most crucial steps for the performance,
as later epochs show mild or negligible improvements.

We also observe the accuracy discrepancy across the three data sets.
Because we keep the same setting for all experiments,
this phenomenon can be best explained by the data's intrinsic property:
the MNIST data is intrinsically more topologically persistent than the other two data sets.
That is, the topological layer is more sensitive to the MNIST data.

\subsection{Learning Speed}

We want to compare the proposed nonparametric method with the parametrized one~\cite{rgabriel_aistat20}.
Because the parameterized topological preprocessing (i.e., sweeping the parameters to obtain the highest accuracy) takes a considerable amount of time\footnote{Parallel computing of this preprocessing stage is possible but we do not discuss it in this paper.},
we restrict this experiment to the MNIST data set at a smaller scale of 400 data points, or MNIST 400.
The network structure (e.g., convolution layers, dropout rates) and other parameters remain the same (e.g., batch size = 100, epoch = 9) as in the previous subsection.

\begin{figure}[htbp]
	\includegraphics[width=\linewidth]{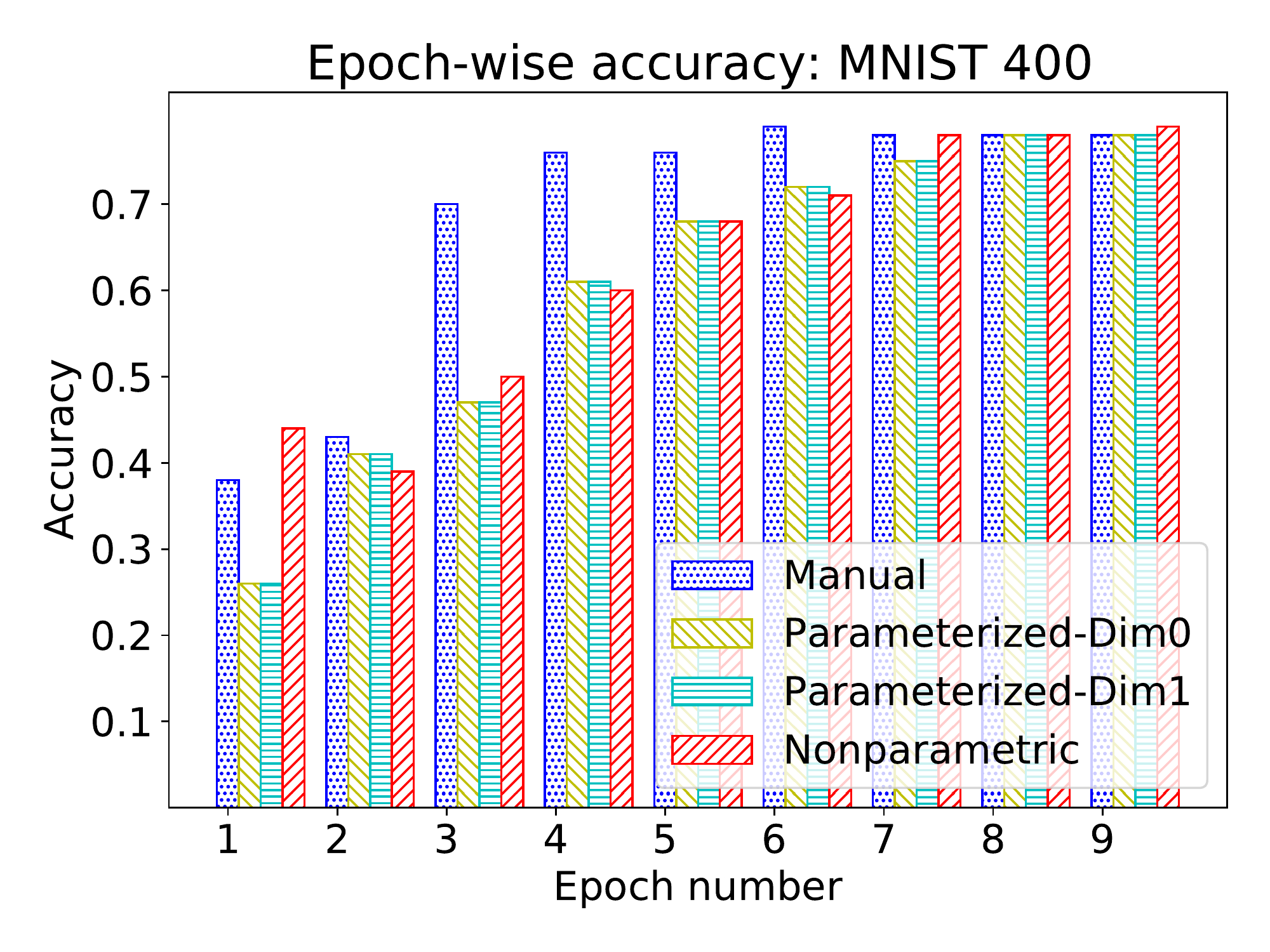}
	\caption{Learning speed of manual optimization, parametrized topologization, and nonparametric topologization.}
	\label{fig:nn_accuracy_mnist}
\end{figure}

Figure~\ref{fig:nn_accuracy_mnist} compares the learning speed of the proposed nonparametric topologization with three baseline methods:
the manual optimization without topologization (starting with the popular \textit{adam}, lr = 0.01),
a parametrized 1-dimensional topologization
and a parametrized 0-dimensional topologization~\cite{rgabriel_aistat20}.
Our experiments show that the nonparametric approach takes seven epochs to reach the optimum, while the manually optimized one only takes four.
Except for the fluctuation at initial epochs, nonparametric and parametrized topologizations exhibit no significant difference in learning speed.
We stress that the main benefit of applying a nonparametric topologization is not for a faster converging rate;
instead, the point of using a nonparametric method is to free the users from the non-systematic procedure of picking the ``optimal'' parameters (trial-and-error, brute force parameter-sweeping) and yet to expect the neural network model
to reach performance on par with that of a handcrafted optimization and/or a parametrized topologization.

To confirm that the nonparametric method indeed offers competitive performance, we enumerate various dimension weights between -8 and +8 for MNIST 400.
Figure~\ref{fig:heatmap_accuracy_mnist} shows that in this parameter space,
the optimal performance can be achieved with parameters at a handful of discrete (integer) combinations, e.g., $(-1,8)$.
Indeed, that subspace of parameters does lead to the performance reported in Figure~\ref{fig:nn_accuracy_mnist}.

\subsection{Temporal-spatial Correlation}
\label{subsec:eval_limit}

\begin{figure}[htbp]
	\includegraphics[width=\linewidth]{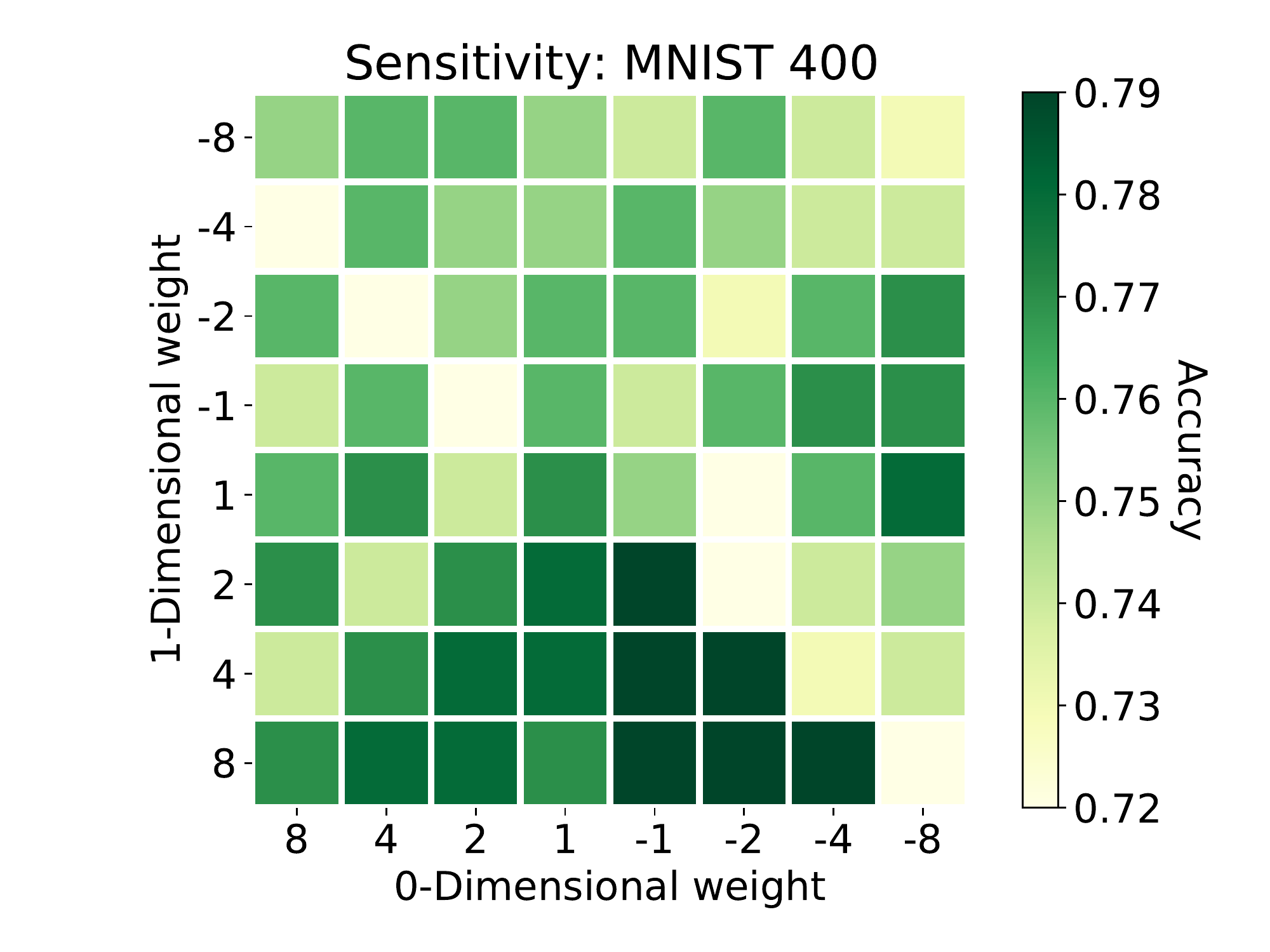}
	\caption{Parameter space of dimension weights for neural network's performance.}
	\label{fig:heatmap_accuracy_mnist}
\end{figure}

The final experiment reports the computation time for the proposed topologization.
While the time overhead of the proposed nonparametric methods is insignificant (20 -- 30 seconds on our testbed) to the neural network training, 
we did observe interesting results when collecting numbers for Figure~\ref{fig:heatmap_accuracy_mnist}.
There seems to be a strong pattern regarding the dimensions of the underlying data set.

\begin{figure}[htbp]
	\includegraphics[width=\linewidth]{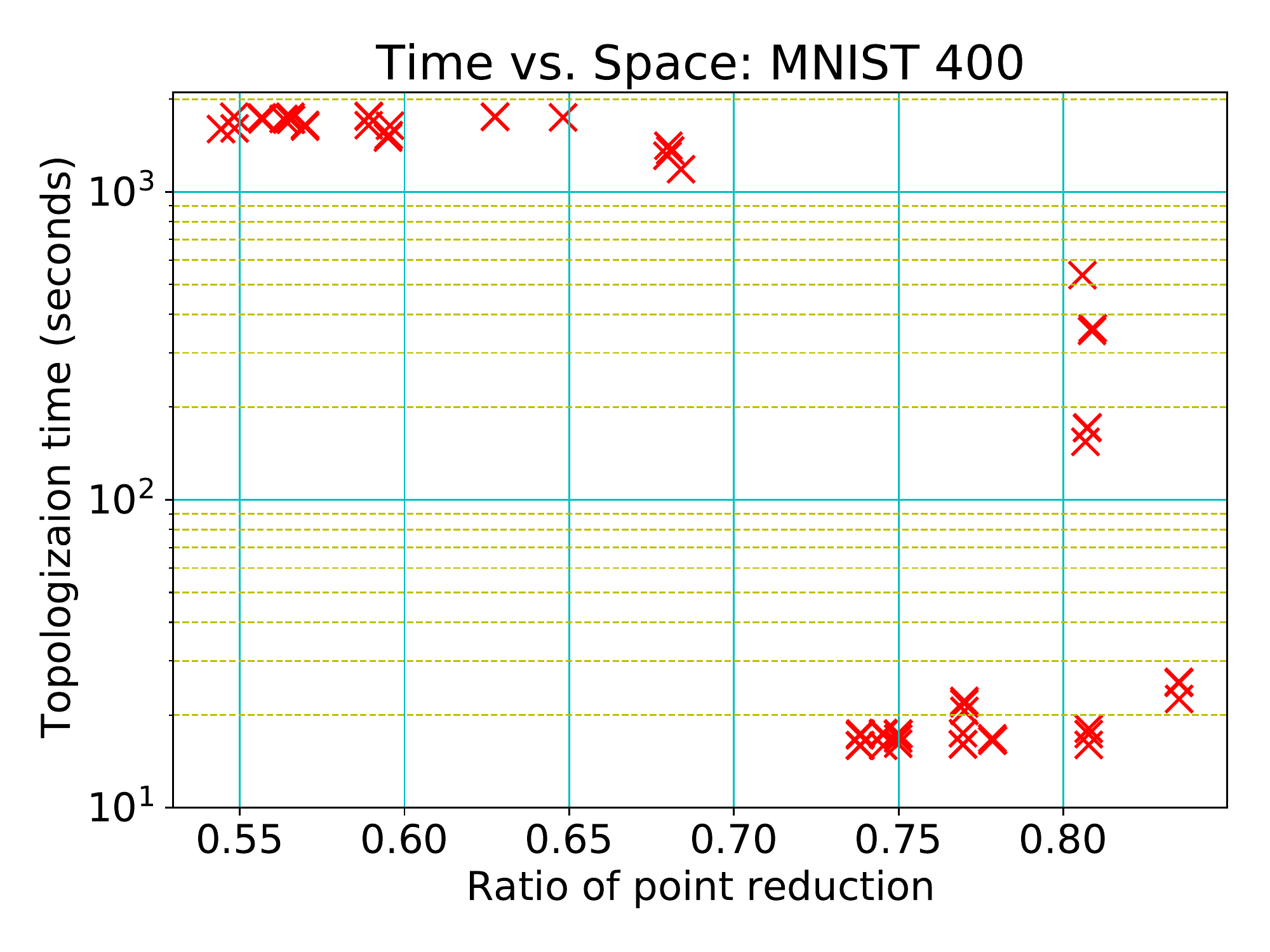}
	\caption{Time-space correlation of topologization.}
	\label{fig:correlation_time_space}
\end{figure}

To investigate the correlation, we conduct 64 topologization experiments with different time-space combinations and reported the results in Figure~\ref{fig:correlation_time_space}.
The correlation seems step-wise (or even binary):
except for four cases (out of 64), about half of the points reside at the top-left corner, and the other half resides at the bottom-right corner.
For MNIST, a ratio of 0.7 seems the cut-off line of these two clusters.
The distance between the two clusters is also interesting:
With only a small difference between space-reduction ratios (e.g., 0.68 vs. 0.73),
the computation time differs in orders of magnitude (e.g., 1,200 seconds vs. 17 seconds).
The root cause of this discrepancy stems from those topological features that are not captured by persistent topology.
How to effectively capture such information is beyond the scope of this paper and will be one of our future research directions.

\section{Additional Related Work}

Adopting topological information in neural networks (esp. machine learning applications) has drawn a lot of research interest.
Much of this line of work focuses on constructing a kernel translating the bar codes into a metric, such as~\cite{rkwitt_nips15,jreinin_cvpr15}.
In addition, some works~\cite{pbub_jmlr15,hadams_jmlr17} show that the kernel can be incorporated with inner products (e.g., in a Banach space).
It should be noted that the above works take the topological characteristics as a \textit{static} feature of the input data.
The first \textit{learnable} layer of topological information for neural networks appeared in~\cite{chofer_nips17}.
The \textit{differentiable} property exhibited in the learned layer then sparkled a new series of works such as leveraging topological information for regularization~\cite{cchen_aistat19}.
Notably, it has been shown that a learnable topological layer can further promote or discount a specific topological feature, as reported in~\cite{rgabriel_aistat20}.
We remark that all of the above works require the users to parametrize the layer,
e.g., through polynomial or exponential functions.

Embedding functional spaces of topological invariant to statistical-learning-based systems has been an active research area for a while.
In~\cite{bsripe_jmlr11}, a distance based on the Gaussian kernel was proposed in the reproducing kernel Hilbert space (RKHS).
Later in~\cite{kmuandet_nips12},
a \textit{support measure machine} (SVM) in the RKHS was proposed for nonparametric inference.
Extending from the Gaussian kernel,
a \textit{weighted Gaussian kernel} for persistent homology was proposed in~\cite{gkusano_icml16}.
In a more general setup, an RKHS embedding was proposed for quantum graphical modelling~\cite{ssrini_nips18}, 
a continuous-time reinforcement-learning framework with RKHS was proposed in~\cite{mohnishi_nips18},
a nonparametric optimization technique in RKHS was proposed for regression applications~\cite{npagliana_nips19},
and a new theory was developed for self-distilling in deep learning in a Hilbert space~\cite{hmobahi_nips20}.
To the best of our knowledge, however, there exists no prior work that systematically constructs a nonparametric learnable topological layer embedded in a Hilbert space. 

\section{Final Remark}
\label{sec:conc}

This paper presents the construction of a nonparametric learnable topological layer embedded in a Hilbert space.
To the best of our knowledge, this is the first learnable topological layer constructed with only inner products,
which makes the layer free of costly parameterization.
We prove that the extended inner products are continuous and the derived loss function (e.g., to be used in gradient-based learning procedure) is differentiable.
We experimentally demonstrate that the proposed nonparametric topological layer achieves competitive performance comparing with the state-of-the-art parametrized layer as well as manual optimization.

Our future work will focus on the time-space correlation exhibited by the nonparametric topological layer.
Specifically, we will turn to other (more advanced) topological tools such as de Rham cohomology,
which may complement the well-studied persistent homology and possibly uncover more insight from the underlying data.

\bibliographystyle{abbrv}
\bibliography{ref_new}

\end{document}